\documentclass[runningheads,a4paper]{llncs}

\usepackage{amssymb}
\usepackage{graphicx}
\usepackage{times}
\usepackage{latexsym}

\usepackage{url}
\usepackage{latexsym}
\usepackage{todonotes}

\usepackage{amsmath}
\usepackage{booktabs}
\usepackage{amssymb}
\usepackage{graphicx}
\usepackage{caption}
\usepackage{url}
\usepackage[outercaption]{sidecap}    

\usepackage[ruled,vlined]{algorithm2e}
\include{pythonlisting}

\DeclareMathOperator*{\argmax}{arg\,max}

\usepackage{gb4e}
\noautomath

\usepackage{array,multirow}

\usepackage{array}
\newcolumntype{L}[1]{>{\raggedright\let\newline\\\arraybackslash\hspace{0pt}}m{#1}}
\newcolumntype{C}[1]{>{\centering\let\newline\\\arraybackslash\hspace{0pt}}m{#1}}
\newcolumntype{R}[1]{>{\raggedleft\let\newline\\\arraybackslash\hspace{0pt}}m{#1}}

\newcommand{\keywords}[1]{\par\addvspace\baselineskip
\noindent\keywordname\enspace\ignorespaces#1}

\begin{document}

\mainmatter  % start of an individual contribution

\title{Addressing Cross-Lingual Word Sense Disambiguation \\ on Low-Density Languages: Application to Persian}

%\titlerunning{}

%\author{Alfred Hofmann%
%\thanks{Please note that the LNCS Editorial assumes that all authors have used
%the western naming convention, with given names preceding surnames. This determines
%the structure of the names in the running heads and the author index.}%
%\and Ursula Barth\and Ingrid Haas\and Frank Holzwarth\and\\
%Anna Kramer\and Leonie Kunz\and Christine Rei\ss\and\\
%Nicole Sator\and Erika Siebert-Cole\and Peter Stra\ss er}
\urldef{\mailsa}\path|[last_name]@ifs.tuwien.ac.at|
\urldef{\mailsb}\path|aduque@lsi.uned.es|

%Andres	Duque	aduque@lsi.uned.es	Spain	Universidad Nacional de Educacion a Distancia (UNED)	

\author{Navid Rekabsaz\inst{1} \and Mihai Lupu\inst{1} \and Allan Hanbury\inst{1} \and Andres Duque\inst{2}}
%
%\authorrunning{Navid Rekabsaz et al.} % abbreviated author list (for running head)
%\tocauthor{
%
\institute{Institute of Software Technology and Interactive Systems\\
TU WIEN\\
A-1040 Vienna, Austria\\ \mailsa
\and
%NLP \& IR Group, Dpto. Lenguajes y Sistemas Informáticos, 
Universidad Nacional de Educación a Distancia (UNED)\\
Madrid 28040, Spain\\ \mailsb
}

%\institute{Springer-Verlag, Computer Science Editorial,\\
%Tiergartenstr. 17, 69121 Heidelberg, Germany\\
%\url{http://www.springer.com/lncs}}


\toctitle{Lecture Notes in Computer Science}
\tocauthor{Authors' Instructions}
\maketitle

\begin{abstract}
We explore the use of unsupervised methods in Cross-Lingual Word Sense Disambiguation (CL-WSD) with the application of English to Persian. Our proposed approach targets the languages with scarce resources (low-density) by exploiting word embedding and semantic similarity of the words in context. We evaluate the approach on a recent evaluation benchmark and compare it with the state-of-the-art unsupervised system (CO-Graph). The results show that our approach outperforms both the standard baseline and the CO-Graph system in both of the task evaluation metrics (\emph{Out-Of-Five} and \emph{Best result}).
\end{abstract}

\section{Introduction}
\label{sec:introduction}
Word Sense Disambiguation (WSD) is the task of automatically selecting the most related sense for a word occurring in a context. It is considered as a main step in the course of approaching language understanding beyond the surface of the words. 

Typically, WSD methods are classified into knowledge-based, supervised, and unsupervised. Knowledge-based approaches use available structured knowledge. Supervised approaches learn a computational model based on large amounts of annotated data. While these two approaches show competitive results in practice, they both have to face the knowledge acquisition bottleneck. This is a particular problem in specific domains or low-density languages. As an alternative, unsupervised approaches address WSD using only information extracted from existing corpora, such as various word co-occurrence indicators.

As two well-known benchmarks for CL-WSD, SemEval-2010~\cite{lefever2010semeval} and SemEval-2013~\cite{lefever2013semeval} provide an evaluation platform for word disambiguation from English to Dutch, German, Italian, Spanish, and French. Recently, Rekabsaz et al.~\cite{rekabsaz2016standard} added the  Persian (Farsi) language to this set by following the CL-WSD SemEval format to create the test collection.

Many participating systems in the SemEval tasks exploit parallel corpora, mainly Europarl~\cite{europarl}, to overcome the knowledge acquisition bottleneck~\cite{lefever2011parasense,rudnick2013hltdi}. However, the approaches used in the tasks are not applicable for many languages and domains due to the scarcity of bilingual corpora. Persian, for instance, suffers from the lack of reliable and comprehensive knowledge resources as well as parallel corpora. In such cases, unsupervised methods based on monolingual corpora (together with bilingual lexicon) are preferable, if not the only available option~\cite{sofianopoulos2012implementing}. For example, Bungum et al.~\cite{bungum2013improving} find the probable translations of a context in the source language and identify the best translation using a language model of the target language. Duque et al.~\cite{duqueco} build a co-occurrence graph in the target language, and test a variety of graph-based algorithms for identifying the best translation match.

In terms of combining Word Sense Disambiguation (WSD) and word embedding, Chen et al.~\cite{chen2014unified} use knowledge-based WSD to identify distinct representations for different senses of the same term. Our approach for CL-WSD is the opposite of this: starting from word embedding representations, it identifies the similarity of the potential translations to the terms in their contexts and choose the translation with the highest semantic similarity to its context.

In order to evaluate our approach, we use the new benchmark of English to Persian CL-WSD~\cite{rekabsaz2016standard}, and compare our approach and the CO-Graph system~\cite{duqueco}, observing the advantages of using word embedding in CL-WSD.

In terms of related work addressing the CL-WSD problem in Persian, Sarrafzadeh et al.~\cite{sarrafzadeh2011} follows a knowledge-based approach by exploiting FarsNet~\cite{shamsfard2010semi}. However, since their evaluation collection is not available, the results are impossible to compare with other possible approaches. 

The remainder of this paper is organized as follows: Section~\ref{sec:methodology} explains our unsupervised approach to English to Persian CL-WSD. We explain the outline of the experiments in Section~\ref{sec:setup}, followed by discussing the result in Section~\ref{sec:experiments}. Finally, the study is concluded in Section~\ref{sec:conclusion}.

\section{Unsupervised CL-WSD Method}
\label{sec:methodology}
Our approach follows the main idea of the Lesk algorithm~\cite{lesk1986automatic}, namely that terms in a given context tend to share a common topic. We use word embedding to compute the semantic similarity between terms. We measure the relatedness of each possible translation of an ambiguous term to all possible translations of the terms in its context (the paragraph given by the task) and select the most similar translation to the context. 

To formulate our CL-WSD approach, let us define $T$ as the list of translation sets for the terms in a context: $T=\{{T_1,T_2,..,T_n} \}$ where $n$ is the number of terms in the context, and $T_i$ is the set of possible translations for the $i^{th}$ term in the context. For each translation $t\in T_i$, we also have $P(t)$---an indicator of how frequent this particular translation is.

Given an embedding model in the target language, we compute the similarity of two translation terms $t$ and $\bar{t}$ using their embedding vectors. However, in some cases the translation $t$ of one term in English may be two or more words in Persian (multi-word term), and since our word embedding model is generally created on words level, we will have more than one vector. Therefore, assuming every term $t$ as a set words $w$, we define a general similarity function between two translation terms as follows:
\begin{equation}
  \textnormal{sim}(t,\bar{t}) = \max_{w \in t, \bar{w} \in \bar{t}}\left(\cos(V_w,V_{\bar{w}})\right) 
  \label{eq:sim}
\end{equation}
where $V_w$ is the vector representation of the word, and $\cos$ is the cosine function. 

%As an example, given the terms `railway' and `coach' and their translations `\<x.t rAh\hspace{0ex}'Ahn>'
%and %`\<wAgn drjh sh>'
%with two and respectively three tokens, the Sim function returns cosine between the vectors of %`\<rAh\hspace{0ex}'Ahn>' and `\<wAgn>'
%(the highest cosine value among the 6 possible combinations). 

Having a definition of similarity between two translation terms, we now move to defining the similarity between a translation candidate of the ambiguous term and the list of translation sets $T$ the terms in the context. We consider two ways to approach it: 

The first, denoted as \emph{RelAgg}, uses the \emph{ContextVec} function to create a vector, representing the translated context terms in the target language. The ContextVec function is defined in Algorithm~\ref{alg:contextVec}.

\begin{algorithm}[h]
\KwIn{translation candidate $t$, and the list of translation sets $T$}
\KwOut{vector representation of the context}

$sumVec\leftarrow []$\;
\For{$ T_i \in  T$}{
$t^*\leftarrow \argmax_{\bar{t} \in T_i}\left(\textnormal{sim}(t, \bar{t})\right)$\;
$maxVec\leftarrow V_{t^*}$\;
$sumVec\leftarrow sumVec+maxVec$\;
}
\textbf{return} $\textnormal{norm}(sumVec)$\;
\caption{ContextVec \label{alg:contextVec}}
\end{algorithm}
The \emph{norm} function in Algorithm~\ref{alg:contextVec} applies the Euclidean norm.

Given the vector representation of the context, RelAgg calculates the cosine between the vector of each translation candidate $t$ to the context vector, multiplied by the probability of the translation candidate $P(t)$, shown as follows:

\begin{equation}
  \textnormal{RelAgg}(t,T)=\cos(V_t,\textnormal{ContextVec}(t,T))P(t)
  \label{eq:relagg}
\end{equation}

The second approach, denoted as \emph{RelGreedy}, searches among all the translation terms in all the sets $T_i$, and returns the value of the most similar translation term to the translation candidate. Similar to RelAgg, the final score is multiplied by the probability of the translation candidate. The RelGreedy approach is defined as follows:
\begin{equation}
  \label{eq:relgreedy}
  \textnormal{RelGreedy}(t,T)=\max_{T_i \in T}\left(\max_{\bar{t} \in T_i} \left(\textnormal{sim}(t,\bar{t}) \right) \right)P(t)
\end{equation}

Finally, given the score of the similarity of each translation candidate $t_i$ to its context using either RelAgg or RelGreedy, we can select the best translation among the candidates, as follows:
\begin{equation}
  \label{eq:bestrel}
  Result=\argmax_{t_i}\left(\textnormal{Rel}^*(t_i,T)\right)
\end{equation}
where $t_i$ is a translation candidate for the term with ambiguity, and $\textnormal{Rel}^*$ is either RelAgg or RelGreedy.

\section{Experiments Setup}
\label{sec:setup}
\paragraph{Resources} Similar to Jadidinejad et al.~\cite{jadidinejad2010evaluation}, we use the PerStem tool~\cite{dehdari2008link} for stemming and TagPer~\cite{seraji2012basic} for POS tagging of Persian language. We create a word2vec SkipGram model on a stemmed corpus of Hamshahri collection~\cite{aleahmad2009hamshahri} with sub-sampling at $t=10^{-4}$, the context windows of 5 terms, epochs of 25, terms count threshold of 5, and dimension of 200.

Beside the monolingual word embedding, a bilingual lexicon is required for our unsupervised CL-WSD approach. While using parallel corpora is considered as a more effective method for creating lexica~\cite{duque2015choosing}, due to the lack of reliable parallel corpora, we have to use a simple English to Persian dictionary. To have it in digital form, we use the online API of one of the existing translation services\footnote{Available in \url{https://github.com/navid-rekabsaz/wsd_persian/tree/master/resources/dictionary}}.

\paragraph{Benchmark} As mentioned before, we use the novel English to Persian CL-WSD collection~\cite{rekabsaz2016standard}, which follows the format of SemEval-2013 test collection. The collection consists of 20 nouns, each with 50 cases (paragraphs) in English where the sense of each noun in its corresponding paragraphs is ambiguous. The aim of the benchmark is to find the correct Persian translations of the ambiguous terms. 

\paragraph{Evaluation} As the official evaluation measure of the SemEval 2013 CL-WSD task~\cite{lefever2013semeval}, we use the F score (harmonic mean of precision and recall), applied in two settings: 
\begin{itemize}
\item \emph{Best Result} (\texttt{Best}), in which a system suggests any number of translations for each target term, and the final score is divided by the number of these translations. 
\item \emph{Out-Of-Five} (\texttt{OOF}) as a more relaxed evaluation setting, in which the system provides up to five different translations, and the best one is selected as the final score.
\end{itemize}

\paragraph{Baselines} The first---STD---is introduced in the SemEval 2013 CL-WSD task as a basic baseline. Similar to the original collection paper, to create the baseline we select the most common and the five most common translations for the \texttt{Best} and \texttt{OOF} settings respectively.

For the second baseline, we evaluate the Persian benchmark on the state-of-the-art unsupervised CL-WSD system, called CO-Graph~\cite{duqueco}. The CO-Graph system offers competitive results in the SemEval 2010 and SemEval 2013 CL-WSD tasks, for all the proposed languages. It outperforms all of the unsupervised participating systems using only monolingual corpora, and even most of the ones which use parallel corpora or knowledge resources. To evaluate the CO-Graph system on the Persian benchmark, we first create the graph using the articles of the Hamshahri collection, each as a document. In the construction of the graph, we only take into account the nouns by POS tagging. After evaluating various algorithms, we find the Dijkstra algorithm together with p-value=$10^{-6}$ as the best performing approach. 

\section{Results and Discussion}
\label{sec:experiments}
\begin{table}[t]
\begin{center}
\caption{Results of F-measure on \texttt{OOF} and \texttt{Best} evaluation settings.}
\begin{tabular}{c l c }
\hline
Setting & Method & F-measure \\
\hline
\multirow{4}{*}{\texttt{OOF}} & RelAgg & \textbf{0.502} \\
& RelGreedy & 0.493 \\
\cline{2-3}
& CO-Graph Dijkstra~\cite{duqueco} & 0.441\\
& STD & 0.418\\
\hline
\hline
\multirow{4}{*}{\texttt{Best}} & RelAgg & \textbf{0.188}  \\
& RelGreedy & 0.183 \\
\cline{2-3}
& CO-Graph Dijkstra~\cite{duqueco} & 0.174\\
& STD & 0.158\\
\hline
\end{tabular}
\label{tbl:results} 
\end{center}
\end{table}
To find the translation for each ambiguous noun, we first apply POS tagging on the English sentences of the SemEval 2013 CL-WSD task and only select the verbs and nouns as the context of the ambiguous terms. We then lemmatize the context terms using WordNetLemmatizer of the NLTK toolkit and find their translations in the bilingual lexicon. Using the word embedding of the translated terms, we finally calculate the relatedness score of each translation candidate to its context using RelAgg and RelGreedy. The translation probability rate in our lexica is used as the $P(t)$ value in Eq.~\ref{eq:relagg} and Eq.~\ref{eq:relgreedy}. 

Table~\ref{tbl:results} shows the F-measure results of RelAgg and RelGreedy as well as the baselines on the \texttt{OOF} and \texttt{Best} evaluation settings. The results for both evaluation settings show that our approach outperforms the standard and the CO-Graph baselines. Comparing the approaches, we observe similar results for the RelAgg and RelGreedy methods, while RelAgg has slightly better performance, specially in the \texttt{OOF} evaluation setting. 

\begin{figure}[t]
\centering
\includegraphics[width=0.8\textwidth]{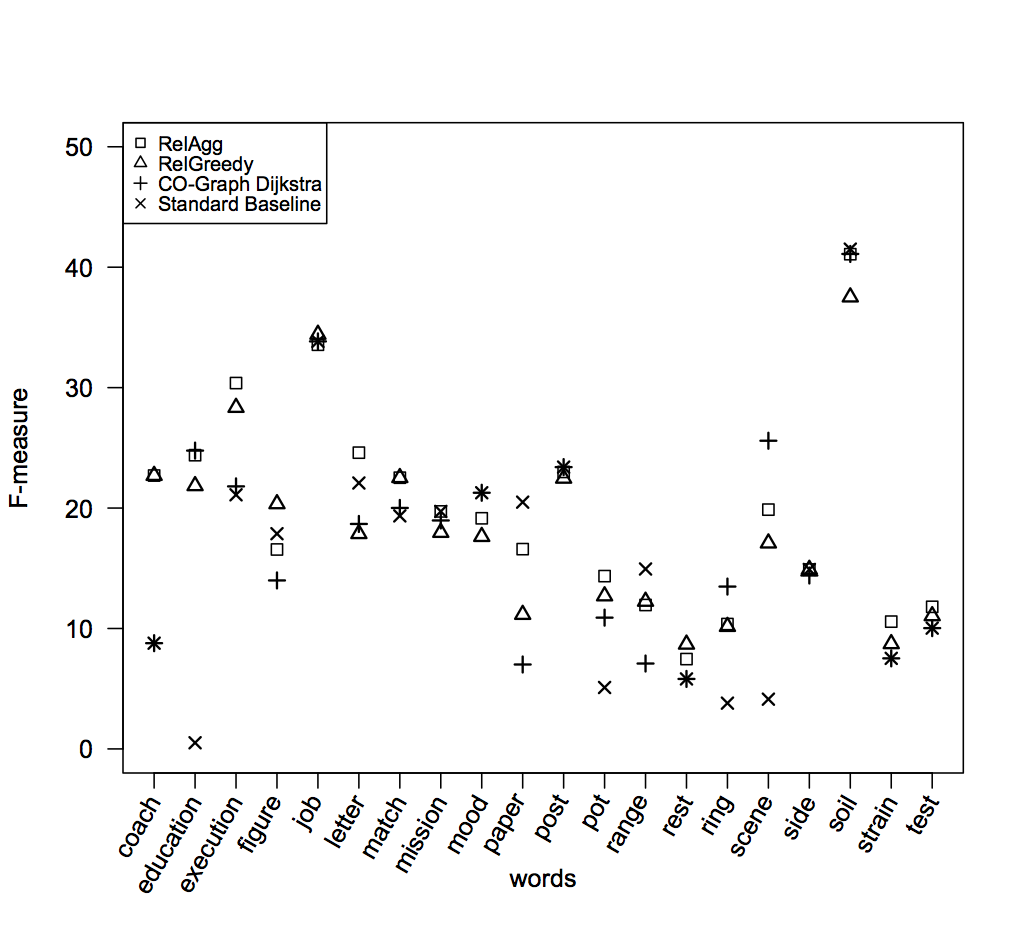}
\caption{F-measure results (multiplied by 100) of the \texttt{Best} evaluation setting, per each term}\label{fig:wordresults-best}
\end{figure}

In Figure~\ref{fig:wordresults-best}, we compare the effectiveness of our methods per each term with the baselines in the \texttt{Best} setting as the more challenging one. The results show that while for most terms, our approach outperforms the standard baseline as well as the CO-Graph system, none of the systems can outperform the standard baseline for the terms `mood' and `side'. Analyzing the results of these terms, we observe that in some sentences, none of the nouns and verbs in the context share any common topic with senses of the ambiguous terms. For example, using only the semantics of the nouns and verbs in the context, the correct sense of `mood' cannot be distinguished in either of the sentences: `it reflected the \textit{mood} of the moment' (state of the feeling) and `a general \textit{mood} in Whitehall' (inclination, tendency) . Similar cases are observed for the term `side': e.g., `both \textit{sides} reaffirmed their commitment' (groups opposing each other) in comparison to `at the  \textit{side} of the cottage' (a position to the left or right of a place). While these examples show the limitations of the context-based methods, the overall results show the ability of word embedding and statistical-based approaches for the CL-WSD tasks, specially in the absence of reliable resources.

\section{Conclusion}
\label{sec:conclusion}
We study the application of word embedding-based methods in unsupervised Cross Language Word Sense Disambiguation (CL-WSD) when translating an English noun, appeared in a paragraph, to Persian. Our semantic approach uses embedding of the candidate translations as well as translated context terms to calculate the semantic similarity of each translation to its context. The proposed approach outperforms both the CO-Graph system---a state-of-the-art system in unsupervised CL-WSD---as well as the standard baseline.

We however observe fundamental limitations of the methods based exclusively on context as bag of words. Despite this fact, the current work offers a possible solution for all languages/domains with scarce knowledge-based or parallel corpora resources, by exploiting the use of a monolingual corpus together with a simple bilingual lexicon.

\bibliographystyle{abbrv}
\vspace{-0.4cm}
\bibliography{refer}
\end{document}